\documentclass[conference]{IEEEtran}
\IEEEoverridecommandlockouts
\usepackage{cite}
\usepackage{amsmath,amssymb,amsfonts,amsthm}
\usepackage{algorithmic}
\usepackage{graphicx}
\usepackage{textcomp}
\usepackage{booktabs}
\usepackage{comment}
\usepackage{xcolor}

\def\BibTeX{{\rm B\kern-.05em{\sc i\kern-.025em b}\kern-.08em
    T\kern-.1667em\lower.7ex\hbox{E}\kern-.125emX}}
    
\DeclareMathOperator*{\argmin}{\arg\!\min}

\theoremstyle{definition}
\newtheorem{definition}{Definition}[]

\begin{document}
	
\onecolumn
\vspace*{.5in}
\noindent \copyright2019 IEEE.  Personal use of this material is permitted.  Permission from IEEE must be obtained for all other uses, in any current or future media, including reprinting/republishing this material for advertising or promotional purposes, creating new collective works, for resale or redistribution to servers or lists, or reuse of any copyrighted component of this work in other works.

\vspace*{0.25in}
\noindent Published in 2019 IEEE International Conference on Fuzzy Systems (FUZZ-IEEE).\\
DOI: 10.1109/FUZZ-IEEE.2019.8858790

\newpage

\twocolumn
\title{Introducing Fuzzy Layers for Deep Learning\\
\thanks{This work was partially supported under the Maneuver in Complex Environments R\&D program to support the U.S. Army ERDC. This effort is also based on work supported by the Defense Threat Reduction Agency/Joint Improvised-Threat Defeat Organization (DTRA/JIDO). Any use of trade names is for descriptive purposes only and does not imply endorsement by the U.S. Government. Permission to publish was granted by Director, Geotechnical and Structures Laboratory, U.S. Army ERDC. Approved for public release; distribution is unlimited.}
}

\author{\IEEEauthorblockN{Stanton R. Price}
\IEEEauthorblockA{\textit{U.S. Army Engineer} \\
\textit{Research and Development Center} \\
Vicksburg, MS USA \\
stantonprice@yahoo.com}
\and
\IEEEauthorblockN{Steven R. Price}
\IEEEauthorblockA{\textit{Department of Electrical Engineering} \\
\textit{Mississippi College}\\
Clinton, MS USA \\
srprice1@mc.edu}
\and
\IEEEauthorblockN{Derek T. Anderson}
\IEEEauthorblockA{\textit{Department of Electrical Engineering}\\
\textit{and Computer Science} \\
\textit{University of Missouri}\\
Columbia, MO, USA \\
andersondt@missouri.edu}
}

\maketitle

\IEEEpubidadjcol

\begin{abstract}
Many state-of-the-art technologies developed in recent years have been influenced by machine learning to some extent.
Most popular at the time of this writing are artificial intelligence methodologies that fall under the umbrella of deep learning.
Deep learning has been shown across many applications to be extremely powerful and capable of handling problems that possess great complexity and difficulty.
In this work, we introduce a new layer to deep learning: the fuzzy layer.
Traditionally, the network architecture of neural networks is composed of an input layer, some combination of hidden layers, and an output layer.
We propose the introduction of fuzzy layers into the deep learning architecture to exploit the powerful aggregation properties expressed through fuzzy methodologies, such as the Choquet and Sugueno fuzzy integrals.
To date, fuzzy approaches taken to deep learning have been through the application of various fusion strategies at the decision level to aggregate outputs from state-of-the-art pre-trained models, e.g., AlexNet, VGG16, GoogLeNet, Inception-v3, ResNet-18, etc.
While these strategies have been shown to improve accuracy performance for image classification tasks, none have explored the use of fuzzified intermediate, or hidden, layers.
Herein, we present a new deep learning strategy that incorporates fuzzy strategies into the deep learning architecture focused on the application of semantic segmentation using per-pixel classification.
Experiments are conducted on a benchmark data set as well as a data set collected via an unmanned aerial system at a U.S. Army test site for the task of automatic road segmentation, and preliminary results are promising.
\end{abstract}


\section{Introduction} 
\textit{Artificial intelligence} (AI) has emerged over the past decade as one of the most promising technologies for advancing mankind in a multitude of ways,
from medicine discovery and disease diagnostics to autonomous vehicles, semantic segmentation, and personal assistants.
For many years there has been a desire to create computer algorithms/machines that are able to replace or assist humans in signal understanding for tasks such as automatic buried explosive hazard detection, vehicle navigation, object recognition, and object tracking.
AI has grown to loosely encompass a number of state-of-the-art technologies across many fields, e.g., pattern recognition, \textit{machine learning} (ML), \textit{neural networks} (NNs), computational intelligence, evolutionary computation, and so on.
Recently, much buzz has surrounded \textit{deep learning} (DL) for its ability to provide desirable results for a number of different applications.

AI has been shown to have great potential for finding optimized solutions to various problems across multiple domains.
This technology has been heavily researched for computer vision applications \cite{bojarski2016end,krizhevsky2012imagenet}, speech translations \cite{bahdanau2014neural,cho2014learning}, and optimization tasks \cite{cortes1995support,kennedy2011particle}.
Herein, the focus is on an extremely relevant and popular branch of AI: deep learning.
DL has achieved much success in recent years on computer vision applications and has benefited from the surge of attention being given to advance its theories to being extremely generalizable for many problems.
Part of DL's recent resurgence is the implementation of its network architecture being well suited for processing on powerful, highly parallelized GPUs.
This has allowed for extremely complex and deep network architectures to be developed that were infeasible to implement on older compute technologies.

Fusion is a powerful technique used to combine information from different sources.
These sources could be different features extracted, decisions, sensor output, etc., as well as different combinations thereof.
Fusion methodologies often strive to improve system performance by combining information in a beneficial way that enables more discriminatory power to the system in some form.
This could be through the realization of more robust features that generalize well from one domain to another.
For the task of classification, fusion could be used to combine multiple decision makers, e.g., classifiers, to improve overall accuracy performance.
Fusion is a very rich and powerful technique that, when implemented appropriately, can lead to major algorithm improvements.
Fusion is commonly associated with fuzzy techniques, as Fuzzy Logic \cite{zadeh1976fuzzy,ross2005fuzzy} naturally lends itself to gracefully considering data with varying degrees of belief.
Ensemble approaches \cite{zhang2012ensemble} are also commonly used for fusion tasks.
Generally, most fusion strategies attempt to properly assign weights that encode the significance, or importance, to the different information sources, and these weights are the driving mechanism behind the fused result.
Historically, weights used when fusing multiple information sources are either human-derived or found via an optimization function/strategy such as group lasso \cite{yuan2006model,meier2008group}.
However, there has been little-to-no research done on utilizing DL for optimizing fusion performance.
Herein, we propose a new strategy to explore the potential benefits of combining DL with fuzzy-based fusion techniques.
Specifically, we introduce \textit{fuzzy layers} to the DL architecture.

The remainder of this work is organized as follows.
In Section \ref{sec:RelatedWork}, related works are presented that explore using fusion strategies to improve the classification performance of the outputs from different state-of-the-art DL models (fusion strategies and DL have been compartmentalized in their utilization).
Fuzzy layers are introduced in Section \ref{sec:Methods} along with the intuition behind this new strategy and their integration into the DL architecture.
Experiments and results are detailed in Section \ref{sec:Exp}, and in Section \ref{sec:Conclusion}, we conclude the paper.

\section{Related Work}
\label{sec:RelatedWork}
As noted previously, DL is becoming the standard approach for classification tasks. However, the performances exhibited by DL classifiers are often the result of exhaustive evaluation of hyperparamenters and various network architectures for the particular data set used in the work. Fusion techniques can help alleviate the comprehensive evaluation by combining the classification outputs from multiple DL classifiers and thus taking advantage of different DL classifier strengths. That is, if the strengths of multiple classifiers can be appropriately fused, then finding the optimal solution may not require finding the ideal parameters and architecture for a particular data set. 

Recently, fusion strategies have been employed that aggregate pre-trained models for improved classification performance. The appropriate fusion strategy largely considers the formats of classifier outputs. Traditionally, a classifier output consists of a definitive label (i.e., hard-decision), and typically, majority voting is the fusion strategy implemented. However, if the classifier can generate soft membership measures (e.g., fuzzy measures), the fusion strategy implemented can vary greatly \cite{ruta2000overview,classfusion,bezdek2001decision,pizzi2010aggregating}. 

Fusion strategies, notably those associated with \textit{fuzzy measures} (FMs), are conventionally applied at the decision level to aggregate outputs and improve performance. For example, DL classifier outputs were fused in \cite{scott2017fusion,scott2018enhanced,anderson2018fuzzy} to improve classification performance in remote sensing applications by deriving FMs for each class and then fusing the measures with the classifiers' outputs through either the Sugeno or \textit{Choquet integral} (ChI) \cite{keller2016fundamentals}. Still, fusion strategies occurring at the input level can also benefit classification performance. 

Rather than attempt to perform fusion at either the output or feature level, it is the attempt of this work to incorporate fusion techniques (utilizing FMs) within the architecture of a DL classification system. While efforts have developed techniques applying fuzzy sets and fuzzy inference systems in NNs, application of fuzzy strategies concerning DL architectures is limited \cite{buckley1994fuzzy}. One recent approach to implementing fuzzy sets in DL evaluated the use of Sugeno fuzzy inference systems as the node in the hidden layer of an NN and therefore could be extended to DL architectures by extending the concept \cite{rajurkar2017developing}.  

\section{Methodology}
\label{sec:Methods}

In this section, we introduce our proposed fuzzy layer to incorporate fuzzy-based fusion approaches directly into the DL architecture.
First, we briefly discuss the problem that is being considered herein: semantic image segmentation using DL.
Semantic segmentation is the process of assigning class labels (e.g., car, tank, road, building, etc.) to each pixel in an image. 
DL for semantic segmentation is most commonly implemented in what can be separated into two parts: (1) standard CNN network with the exception of an output layer and (2) an up-sampling CNN network on the back half that produces a per-pixel class label output.
Zeiler et al. introduced deconvolution networks in \cite{zeiler2014visualizing} for the task of visualizing learned CNNs to help bridge the gap of understanding what a CNN has learned.
Therein, Zeiler et al. defined a deconvolution network that attempts to reconstruct a feature map that identifies what a CNN has learned through unpooling, rectification, and filtering. 
In \cite{noh2015learning}, Noh et al. modified this approach for semantic segmentation by using DL to learn the up-sampling, or deconvolution, filters rather than inverting (via transposing) the corresponding filters learned in the convolution network.
Herein, we implement a similar approach as Noh et al., utilizing DL strategies to learn the up-sampling filters rather than performing true deconvolution to reconstruct the feature maps at each layer.
Additionally, this work is focused strictly on road segmentation, i.e., each pixel is labeled as either road or non-road.
Representing the architecture of our learned model as $f(x,\gamma)$, where $\gamma$ represents the parameters that are learned by the network such that the error for an example $x_i$ given its ground-truth $y_i$ is minimized and can be described as
\begin{equation}
    \hat{\gamma}=\argmin_\gamma \sum_{i=1}^{N}(L(f(x_i,\gamma),y_i), 
\end{equation}

\noindent where $N$ is the training data set and $L$ is the $softmax$ (cross-entropy) loss.

As this paper is focused on the introduction of a new fuzzy layer that can be injected directly into the DL architecture, a defined network architecture is not presented.
Rather, we explore different use cases of the fuzzy layers at different points throughout the network architecture.
For comparison, the fuzzy layers are utilized either in the down-sampling (convolution network), up-sampling (``deconvolution" network), or both sections of the semantic segmentation network.
To maintain consistency in our exploration, a template network architecture was used such that the only change in the network architecture was the inclusion or removal of one or more fuzzy layers.
The details of the architecture template used are given in Table \ref{tab:archTemplate}, with `*' denoting the points in which a fuzzy layer \textit{might} be included in the architecture.
\textbf{Note:} it is \textit{not required} that a fuzzy layer be incorporated after a \textit{rectified linear unit} (ReLU); this occurs in the results presented herein to maintain more consistency across experiments in this exploratory work.
It would have been equally valid to implement a fuzzy layer after any convolution or pooling layer (referencing layers utilized in this architecture).
The best way(s) to implement fuzzy layers within the DL architecture is an open question and one that requires additional research. 
Technically, a fuzzy layer can be implemented anywhere in the DL architecture as long as it follows the input layer and precedes the output layer.

\begin{table}[]
\caption{Template architecture in detail. The `*' represents locations in the architecture that a fuzzy layer might be included herein. This is not a restriction. $N_{f}$ and $N_{cl}$ represent the number of fused outputs at that layer and the number of classes, respectively.}
\label{tab:archTemplate}
\centering
\begin{tabular}{@{}lccl@{}}
\toprule
\multicolumn{1}{c}{\textbf{Name}} & \textbf{Kernel Size} & \textbf{Stride} & \textbf{Output Size}       \\ \midrule
input\_data                       & -                    & -               & $512\times512\times3$      \\
conv1\_1                          & $5\times5$           & 1               & $512\times512\times64$     \\
conv1\_2                          & $5\times5$           & 1               & $512\times512\times64$     \\
relu1                             & -                    & -               & $512\times512\times64$     \\
*                                 & -                    & -               & $512\times512\times N_{f}$ \\
pool1                             & $2\times2$           & 2               & $256\times256\times64$     \\
conv2\_1                          & $5\times5$           & 1               & $256\times256\times64$     \\
relu2                             & -                    & -               & $256\times256\times64$     \\
*                                 & -                    & -               & $256\times256\times N_{f}$ \\
pool2                             & $2\times2$           & 2               & $128\times128\times64$     \\
conv3\_1                          & $5\times5$           & 1               & $128\times128\times64$     \\
relu3                             & -                    & -               & $128\times128\times64$     \\
*                                 & -                    & -               & $128\times128\times N_{f}$ \\
up-conv1                          & $6\times6$           & 2               & $256\times256\times30$     \\
relu4                             & -                    & -               & $256\times256\times30$     \\
*                                 & -                    & -               & $256\times256\times N_{f}$ \\
up-conv2                          & $6\times6$           & 2               & $512\times512\times30$     \\
relu5                             & -                    & -               & $512\times512\times30$     \\
*                                 & -                    & -               & $512\times512\times N_{f}$ \\
output                            & -                    & -               & $512\times512\times N_{cl}$ \\ \bottomrule
\end{tabular}
\end{table}

\subsection{Fuzzy Layer}
Theoretically, the fuzzy layer can encompass any fuzzy aggregation strategy desired to be utilized.
Herein, we focus on the Choquet integral as the means for fusion.
Let $X = \{x_1,\ldots,x_N\}$ be $N$ sources, e.g., sensor, human, or algorithm.
In general, an aggregation function is a mapping of data from our $N$ sources, denoted by $h(x_i)\in\mathcal{R}$, to data, $f({h(x_1),\dots,h(x_N)},\Theta)\in \mathcal{R}$, where $\Theta$ are the parameters of $f$.
The ChI is a nonlinear aggregation function parameterized by the FM.
FMs are often used to encode the (possibly subjective) worth of different subsets of information sources.
Thus, the ChI parameterized by the FM provides a way to combine the information encoded in the FM with the (objective) evidence or support of some hypothesis, e.g., sensor data, algorithm outputs, expert opinions, etc.
The FM and ChI are defined as follows.

\theoremstyle{definition}
\begin{definition}{\textbf{(Fuzzy Measure)}}
For a finite set of $N$ information sources, $X$, the FM is a set-valued function $g:2^X\rightarrow[0,1]$ with the following conditions:
\begin{enumerate}
    \item (Boundary Conditions) $g(\emptyset)=0$ and $g(X)=1$
    \item (Monotonicity) If $A$, $B \subseteq X$ with $A \subseteq B$, then $g(A) \leq g(B)$.
\end{enumerate}
Note, if $X$ is an infinite set, there is a third condition guaranteeing continuity.
\end{definition}

\begin{figure*}
    \centering
    \includegraphics[scale=0.85]{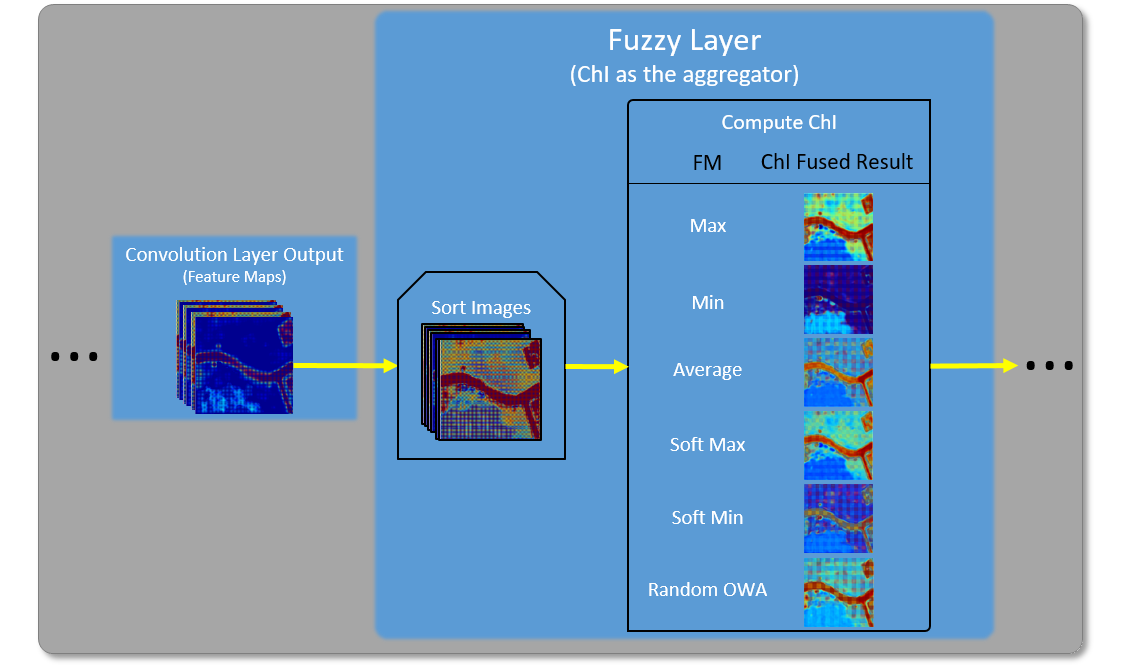}
    \caption{Illustration of the fuzzy layer. In this example, the layer feeding into the fuzzy layer is a convolution layer. The feature maps are passed as inputs to the fuzzy layer where they are then sorted, as required for the ChI, based on some metric (entropy used herein). The ChI is computed for six different FMs, producing six ChI fused resultant images. These six images are then passed on to the next layer in the network.}
    \label{fig:fuzzyLayerEx}
\end{figure*}

\theoremstyle{definition}
\begin{definition}{\textbf{(Choquet Integral)}}
For a finite set of $N$ information sources, $X$, FM $g$, and partial support function $h:X\rightarrow [0,1]$, the ChI is
\begin{equation}
    \int h \circ g = \sum_{i=1}^{N}w_ih(x_{\pi(i)}),
\end{equation}
\noindent where $w_i=(G_{\pi(i)}-G_{\pi(i-1)})$, $G_{(i)}=g(\{x_{\pi(1)},\ldots,x_{\pi(i)}\})$, $G_{\pi(0)}=0$, $h(x_i)$ is the strength in the hypothesis from source $x_i$, and $\pi(i)$ is a sorting on $X$ such that $h(x_{\pi(1)})\geq \ldots \geq h(x_{\pi(N)})$

The FM can be obtained in a number of ways: human defined, quadratic program, learning algorithm, S-Decomposable measure (e.g., Sugeno $\lambda$-fuzzy measure), etc.
Herein, we define the FM to be five known OWA operators and one random (but valid) OWA operator.
Specifically, the more well known operators used are $\max$, $\min$, average, soft $\max$, and soft $\min$.
The top 5 sources (i.e., convolution/deconvolution filter outputs) were sorted based on their entropy value and fused via the ChI.
Therefore, the fuzzy layer accepts the output from the previous layer as its input, sorts the images (sources) by some metric (entropy used herein), and performs the ChI for each of the defined FMs resulting in six fused outputs (we have six different FMs) that are passed on to the next layer in the network.
An example of a potential fuzzy layer implementing the ChI as its aggregation method is shown in Figure \ref{fig:fuzzyLayerEx}.
\end{definition}

\subsection{Why Have Fuzzy Layers?}
As the architecture of DL continues to grow more complex, there is a need to help alleviate the ill-conditioning that is prone to occur during learning due to the weights approaching zero.
Additionally, a well-known problem when training deep networks is the internal-covariate-shift problem \cite{ioffe2015batch}.
This results in difficulty to optimize the network due to the input distributions changing at each layer over iterations during training with the changes in distribution being amplified through propagation across layers.
While there are other approaches that seek to help with this (e.g., batch normalization \cite{ioffe2015batch}), fusion poses itself as a viable solution to aiding with this problem.
One example of the potential benefit of this is fusion can take 10s, 100s, etc. of inputs (outputs of previous layers) and condense that information into a fraction of images.
For example, if the output of a convolution, ReLU, or pooling layer had 256 feature maps, a fuzzy layer could be utilized to fuse these 256 feature maps down to some arbitrary reduced number of feature maps, e.g., 30, that capture relevant information in unique ways from all, or some subset of the 256 feature maps (dependent on the FMs used as well as the metric used for sorting).
Thus, this alone has two potential major benefits: (1) reduced model complexity and (2) improving the utilization of the information learned at the previous layer in the network.

\section{Experiments \& Results}
\label{sec:Exp}

This section first describes the dataset and implementation details.
Next, we present and analyze the results for various network configurations as we investigate the implementation of fuzzy layers.

\subsection{Dataset}
\begin{figure}
    \centering
    \includegraphics{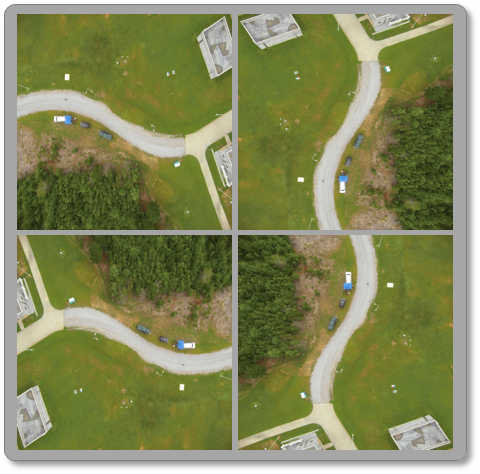}
    \caption{Example of data set augmentation used (image rotation). Starting at the top left and going clockwise: 0\textdegree, 90\textdegree, 180\textdegree, 270\textdegree.}
    \label{fig:SampleImsRotated}
\end{figure}

\begin{figure}
    \centering
    \includegraphics[scale=0.85]{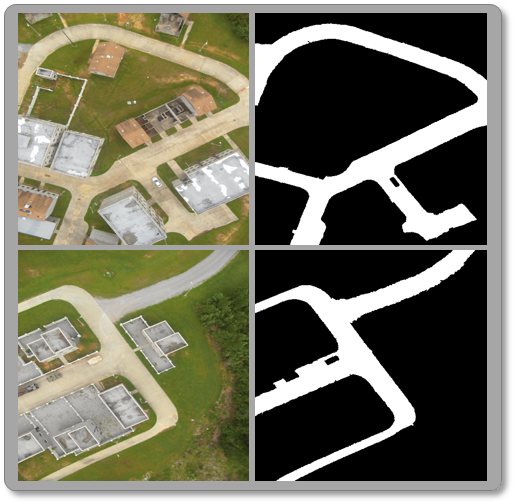}
    \caption{Road masks shown (right column) for two sample images.}
    \label{fig:SampleRoadMasks}
\end{figure}

The dataset was collected from an UAS with a mounted MAPIR Survey2 RGB camera.
Specifically, the sensor used is a Sony Exmor IMX206 16MP RGB sensor, which produces a 24-bit 4,608$\times$3,456 pixel RGB image.
The UAS was flown at an altitude of approximately 60 meters from the ground.
The dataset was captured by flying the UAS in a grid-like pattern over an area of interest at a U.S. Army test site.
The dataset used in this work comes from a single flight over this area, which contains 252 images, 20 of which were selected as training data.
The imagery was scaled to size (i.e., 512$\times$512 pixels) using bilinear interpolation to make them more digestible by the DL algorithms implemented on the computer system used herein.
As is common with training for DL algorithms (and ML algorithms in general), data augmentation strategies are employed in part to increase the amount of training data available during learning, but also to lead to more robust solutions.
Herein, each image from the training data set is rotated 90\textdegree, 180\textdegree, and 270\textdegree \ to provide a total of 80 images used for training (example shown in Figure \ref{fig:SampleImsRotated}).
Finally, the image road mask for each of the 252 instances were annotated by hand (see Figure \ref{fig:SampleRoadMasks}). 

\subsection{Implementation Details}
We based the template network architecture shown in Table \ref{tab:archTemplate} on the VGG-16 framework \cite{simonyan2014very}.
There are modifications to the number of convolution layers and filters used throughout the network; however, the VGG-16 framework served as the motivation behind the defined template architecture implemented.
Initially, we implemented the standard stochastic gradient descent with momentum for optimization but achieved poor results.
The Adam algorithm \cite{kingma2014adam} provided the best optimization performance on this dataset and network design and was used for all experiments reported, with the initial learning rate, gradient decay rate, and squared gradient decay weight set to 0.1, 0.9, and 0.999, respectively.
Dropout \cite{srivastava2014dropout} is used after pooling with a dropout rate of $50\%$.

\subsection{Evaluation}

To measure network classification accuracy performance, we utilize an evaluation protocol that is based on \textit{Intersection over Union} (IoU) between the ground-truth and predicted segmentation.
We report the mean and standard deviation of the IoU scores for all test images for each approach investigated.
For clarity, we denote the different experiments (i.e., different architecture configurations) as follows
\begin{itemize}
    \item \textbf{baseline--} no fuzzy layers;
    \item \textbf{conv-FLs--} fuzzy layers are implemented after `relu1', `relu2', and `relu3' in the convolution network (down-sampling half);
    \item \textbf{deconv-FLs--} fuzzy layers are implemented after `relu4' and `relu5' in the ``deconvolution" network (up-sampling half);
    \item  \textbf{conv-FLs+deconv-FLs--} fuzzy layers are implemented after every ReLU.
\end{itemize}
The quantitative results for these different architectures are presented in Table \ref{tab:results}.

\begin{table}[]
\centering
\caption{Evaluation results on test dataset for road segmentation.}
\label{tab:results}
\begin{tabular}{@{}lcc@{}}
\toprule
Method & Mean & Std. Dev. \\ \midrule
baseline & 78.22\%  & \textbf{12.3\%} \\
conv-FLs & 62.43\% & 21.5\% \\
deconv-FLs & \textbf{80.79\%} & 14.8\% \\
conv-FLs+deconv-FLs & 68.76\% & 20.7\% \\ \bottomrule
\end{tabular}
\end{table}

\begin{figure}
    \centering
    \includegraphics[width=.48\textwidth]{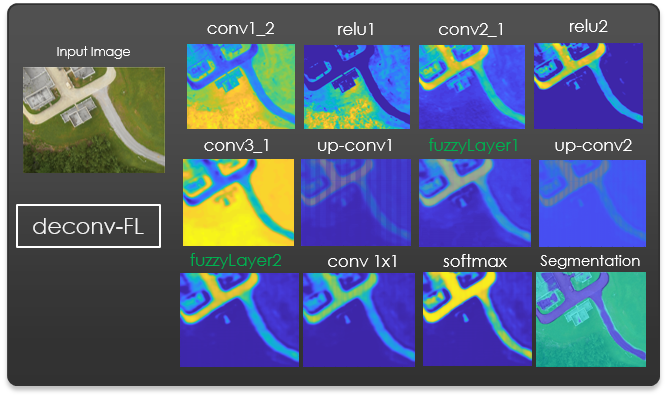}
    \caption{Example feature maps and final segmentation for a randomly selected image. Also, the feature maps were randomly chosen at each layer.}
    \label{fig:deconv_FuzzyLayers_NetworkEx}
\end{figure}

From these preliminary results, we see that the inclusion of fuzzy layers shows promise for improving DL performance (in terms of accuracy).
In particular, these results indicate that fuzzy layers are better utilized in the deconvolution phase of the architecture.
Example feature maps randomly selected from one instance at each layer (the ReLU output is ommited in the deconvolution network for compactness) are shown in Figure \ref{fig:deconv_FuzzyLayers_NetworkEx}.
Looking specifically at the feature maps denoted as `fuzzyLayer1' and `fuzzyLayer2', we see evidence of the fuzzy layers' aggregation strategy accumulating evidence of road information.
We note that `deconv-FLs' only performs approximately $2\%$ better than the baseline method, while having a slightly higher standard deviation.
Nevertheless, this helps show fuzzy layers potential of improving classification performance.
It is our conjecture that, for this problem, applying the fuzzy layers during the convolution stage (results shown as `conv-FLs') results in the loss of too much information from prior layers (after each ReLU, we summarize 64 filters down to 6-- this is likely too extreme for such an early stage of learning).
Hence, we see a noticeable drop in performance for both experiments that include fuzzy layers during the convolution stage (`conv-FLs' and `conv-FLs+deconv-FLs').
However, there are a number of factors involved that could lead to improved performance during the convolution phase, e.g., increased number of FMs, perhaps a different metric for sorting should be used, different fuzzy aggregation method, etc.
It should also be noted that the inclusion of the fuzzy layers had minimal impact on training time (total training time increased by seconds to a few minutes at most).

\section{Conclusion}
\label{sec:Conclusion}
We proposed a new layer to be used for DL: the fuzzy layer.
The proposed fuzzy layer is extremely flexible, capable of implementing any fuzzy aggregation method desired, as well as capable of being included anywhere in the network architecture, depending on the desired behavior of the fuzzy layer.
This work was focused on the introduction and early exploration of the fuzzy layer, and additional research is needed to further advance the fuzzy layer for DL.
For example, future work should consider investigating the metric used for sorting the information sources and its effect on accuracy performance.
Future work is planned to investigate how the FM should be defined for aggregating via fuzzy integrals.
Additionally, where are the fuzzy layers best utilized in the network architecture (problem dependent; however, can a general guidance be developed)?
These are but a few questions that need to be addressed for the fuzzy layer and its implementation.

\bibliographystyle{IEEEtran}
\bibliography{Refs}

\begin{thebibliography}{10}
\providecommand{\url}[1]{#1}
\csname url@samestyle\endcsname
\providecommand{\newblock}{\relax}
\providecommand{\bibinfo}[2]{#2}
\providecommand{\BIBentrySTDinterwordspacing}{\spaceskip=0pt\relax}
\providecommand{\BIBentryALTinterwordstretchfactor}{4}
\providecommand{\BIBentryALTinterwordspacing}{\spaceskip=\fontdimen2\font plus
\BIBentryALTinterwordstretchfactor\fontdimen3\font minus
  \fontdimen4\font\relax}
\providecommand{\BIBforeignlanguage}[2]{{%
\expandafter\ifx\csname l@#1\endcsname\relax
\typeout{** WARNING: IEEEtran.bst: No hyphenation pattern has been}%
\typeout{** loaded for the language `#1'. Using the pattern for}%
\typeout{** the default language instead.}%
\else
\language=\csname l@#1\endcsname
\fi
#2}}
\providecommand{\BIBdecl}{\relax}
\BIBdecl

\bibitem{bojarski2016end}
M.~Bojarski, D.~Del~Testa, D.~Dworakowski, B.~Firner, B.~Flepp, P.~Goyal, L.~D.
  Jackel, M.~Monfort, U.~Muller, J.~Zhang \emph{et~al.}, ``End to end learning
  for self-driving cars,'' \emph{arXiv preprint arXiv:1604.07316}, 2016.

\bibitem{krizhevsky2012imagenet}
A.~Krizhevsky, I.~Sutskever, and G.~E. Hinton, ``Imagenet classification with
  deep convolutional neural networks,'' in \emph{Advances in neural information
  processing systems}, 2012, pp. 1097--1105.

\bibitem{bahdanau2014neural}
D.~Bahdanau, K.~Cho, and Y.~Bengio, ``Neural machine translation by jointly
  learning to align and translate,'' \emph{arXiv preprint arXiv:1409.0473},
  2014.

\bibitem{cho2014learning}
K.~Cho, B.~Van~Merri{\"e}nboer, C.~Gulcehre, D.~Bahdanau, F.~Bougares,
  H.~Schwenk, and Y.~Bengio, ``Learning phrase representations using rnn
  encoder-decoder for statistical machine translation,'' \emph{arXiv preprint
  arXiv:1406.1078}, 2014.

\bibitem{cortes1995support}
C.~Cortes and V.~Vapnik, ``Support-vector networks,'' \emph{Machine learning},
  vol.~20, no.~3, pp. 273--297, 1995.

\bibitem{kennedy2011particle}
J.~Kennedy, ``Particle swarm optimization,'' in \emph{Encyclopedia of machine
  learning}.\hskip 1em plus 0.5em minus 0.4em\relax Springer, 2011, pp.
  760--766.

\bibitem{zadeh1976fuzzy}
L.~A. Zadeh, ``A fuzzy-algorithmic approach to the definition of complex or
  imprecise concepts,'' in \emph{Systems Theory in the Social Sciences}.\hskip
  1em plus 0.5em minus 0.4em\relax Springer, 1976, pp. 202--282.

\bibitem{ross2005fuzzy}
T.~J. Ross, \emph{Fuzzy logic with engineering applications}.\hskip 1em plus
  0.5em minus 0.4em\relax John Wiley \& Sons, 2005.

\bibitem{zhang2012ensemble}
C.~Zhang and Y.~Ma, \emph{Ensemble machine learning: methods and
  applications}.\hskip 1em plus 0.5em minus 0.4em\relax Springer, 2012.

\bibitem{yuan2006model}
M.~Yuan and Y.~Lin, ``Model selection and estimation in regression with grouped
  variables,'' \emph{Journal of the Royal Statistical Society: Series B
  (Statistical Methodology)}, vol.~68, no.~1, pp. 49--67, 2006.

\bibitem{meier2008group}
L.~Meier, S.~Van De~Geer, and P.~B{\"u}hlmann, ``The group lasso for logistic
  regression,'' \emph{Journal of the Royal Statistical Society: Series B
  (Statistical Methodology)}, vol.~70, no.~1, pp. 53--71, 2008.

\bibitem{ruta2000overview}
D.~Ruta and B.~Gabrys, ``An overview of classifier fusion methods,''
  \emph{Computing and Information systems}, vol.~7, no.~1, pp. 1--10, 2000.

\bibitem{classfusion}
Z.~Liu, Q.~Pan, J.~Dezert, J.~Han, and Y.~He, ``Classifier fusion with
  contextual reliability evaluation,'' \emph{IEEE Transactions on Cybernetics},
  vol.~48, no.~5, pp. 1605--1618, May 2018.

\bibitem{bezdek2001decision}
L.~I. Kuncheva, J.~C. Bezdek, and R.~P. Duin, ``Decision templates for multiple
  classifier fusion: an experimental comparison,'' \emph{Pattern recognition},
  vol.~34, no.~2, pp. 299--314, 2001.

\bibitem{pizzi2010aggregating}
N.~J. Pizzi and W.~Pedrycz, ``Aggregating multiple classification results using
  fuzzy integration and stochastic feature selection,'' \emph{International
  Journal of Approximate Reasoning}, vol.~51, no.~8, pp. 883--894, 2010.

\bibitem{scott2017fusion}
G.~J. Scott, R.~A. Marcum, C.~H. Davis, and T.~W. Nivin, ``Fusion of deep
  convolutional neural networks for land cover classification of
  high-resolution imagery,'' \emph{IEEE Geoscience and Remote Sensing Letters},
  vol.~14, no.~9, pp. 1638--1642, 2017.

\bibitem{scott2018enhanced}
G.~J. Scott, K.~C. Hagan, R.~A. Marcum, J.~A. Hurt, D.~T. Anderson, and C.~H.
  Davis, ``Enhanced fusion of deep neural networks for classification of
  benchmark high-resolution image data sets,'' \emph{IEEE Geoscience and Remote
  Sensing Letters}, vol.~15, no.~9, pp. 1451--1455, 2018.

\bibitem{anderson2018fuzzy}
D.~T. Anderson, G.~J. Scott, M.~A. Islam, B.~Murray, and R.~Marcum, ``Fuzzy
  choquet integration of deep convolutional neural networks for remote
  sensing,'' in \emph{Computational Intelligence for Pattern
  Recognition}.\hskip 1em plus 0.5em minus 0.4em\relax Springer, 2018, pp.
  1--28.

\bibitem{keller2016fundamentals}
J.~M. Keller, D.~B. Fogel, and D.~Liu, \emph{Fundamentals of computational
  intelligence: neural networks, fuzzy systems, and evolutionary
  computation}.\hskip 1em plus 0.5em minus 0.4em\relax John Wiley \& Sons,
  2016.

\bibitem{buckley1994fuzzy}
J.~J. Buckley and Y.~Hayashi, ``Fuzzy neural networks: A survey,'' \emph{Fuzzy
  sets and systems}, vol.~66, no.~1, pp. 1--13, 1994.

\bibitem{rajurkar2017developing}
S.~Rajurkar and N.~K. Verma, ``Developing deep fuzzy network with takagi sugeno
  fuzzy inference system,'' in \emph{Fuzzy Systems (FUZZ-IEEE), 2017 IEEE
  International Conference on}.\hskip 1em plus 0.5em minus 0.4em\relax IEEE,
  2017, pp. 1--6.

\bibitem{zeiler2014visualizing}
M.~D. Zeiler and R.~Fergus, ``Visualizing and understanding convolutional
  networks,'' in \emph{European conference on computer vision}.\hskip 1em plus
  0.5em minus 0.4em\relax Springer, 2014, pp. 818--833.

\bibitem{noh2015learning}
H.~Noh, S.~Hong, and B.~Han, ``Learning deconvolution network for semantic
  segmentation,'' in \emph{Proceedings of the IEEE international conference on
  computer vision}, 2015, pp. 1520--1528.

\bibitem{ioffe2015batch}
S.~Ioffe and C.~Szegedy, ``Batch normalization: Accelerating deep network
  training by reducing internal covariate shift,'' \emph{arXiv preprint
  arXiv:1502.03167}, 2015.

\bibitem{simonyan2014very}
K.~Simonyan and A.~Zisserman, ``Very deep convolutional networks for
  large-scale image recognition,'' \emph{arXiv preprint arXiv:1409.1556}, 2014.

\bibitem{kingma2014adam}
D.~P. Kingma and J.~Ba, ``Adam: A method for stochastic optimization,''
  \emph{arXiv preprint arXiv:1412.6980}, 2014.

\bibitem{srivastava2014dropout}
N.~Srivastava, G.~Hinton, A.~Krizhevsky, I.~Sutskever, and R.~Salakhutdinov,
  ``Dropout: a simple way to prevent neural networks from overfitting,''
  \emph{The Journal of Machine Learning Research}, vol.~15, no.~1, pp.
  1929--1958, 2014.

\end{thebibliography}

\end{document}